\documentclass[letterpaper]{article} 
\usepackage{aaai24}  
\usepackage{times}  
\usepackage{helvet}  
\usepackage{courier}  
\usepackage[hyphens]{url}  
\usepackage{graphicx} 
\def\UrlFont{\rm}  
\usepackage{natbib}  
\usepackage{caption} 
\usepackage{algorithm}
\usepackage{algorithmic}
\usepackage{amsfonts}

\usepackage{newfloat}
\usepackage{listings}
\DeclareCaptionStyle{ruled}{labelfont=normalfont,labelsep=colon,strut=off} 
\floatstyle{ruled}
\newfloat{listing}{tb}{lst}{}
\floatname{listing}{Listing}
\usepackage{times}
\usepackage{soul}
\usepackage{url}
\usepackage[utf8]{inputenc}
\usepackage{graphicx}
\usepackage{amsmath}
\usepackage{amsthm}
\usepackage{booktabs}
\usepackage{algorithm}
\usepackage{algorithmic}
\usepackage{subcaption}
\usepackage{tabularx}
\title{Toward More Generalized Malicious URL Detection Models}

\author {
    Yun-Da Tsai,
    Cayon Liow,
    Yin Sheng Siang,
    Shou-De Lin\\
}
\affiliations {
    National Taiwan University \\
    f08946007@csie.ntu.edu.tw, cayon.1318.96@hotmail.com\\
    b06902103@csie.ntu.edu.tw, sdlin@csie.ntu.edu.tw\\
}

\usepackage{bibentry}

\begin{document}

\maketitle

\begin{abstract}
This paper reveals a data bias issue that can profoundly hinder the performance of machine learning models in malicious URL detection. We describe how such bias can be diagnosed using interpretable machine learning techniques and further argue that such biases naturally exist in the real world security data for training a classification model. To counteract these challenges, we propose a debiased training strategy that can be applied to most deep-learning based models to alleviate the negative effects of the biased features. The solution is based on the technique of adversarial training to train deep neural networks learning invariant embedding from biased data. Through extensive experimentation, we substantiate that our innovative strategy fosters superior generalization capabilities across both CNN-based and RNN-based detection models. The findings presented in this work not only expose a latent issue in the field but also provide an actionable remedy, marking a significant step forward in the pursuit of more reliable and robust malicious URL detection.
\end{abstract}

\section{Introduction}
Machine learning techniques have become indispensable in security-related applications like malware detection, URL reputation classification, and intrusion detection to combat the growing variety of Internet attacks~\cite{gandotra2014malware,tsai2021toward}. Recent end-to-end solutions utilizing deep learning models have achieved state-of-the-art performance, reporting cross-validation AUC scores above 98\% when trained on specific datasets.

However, our study reveals a significant vulnerability in these cutting-edge detection models: a serious deterioration in performance when applied to real-world daily queries. Through extensive testing of three leading malware detection models (URLNet, Malconv, and LSTM) using the ISCX-URL-2016 dataset, we observed a drop in performance at test time.
The results in Table~\ref{tab:bad_performace_a} show at least a 30.2\% drop in performance when evaluating with open-domain data - large daily queries obtained from VirusTotal. We further evaluated the models trained on VirusTotal data from 2013-2017 and also tested on VirusTotal from 2018-2019.
The results in Table~\ref{tab:bad_performace_b} also showed at least a 30\% drop in performance which indicates an inability to generalize to changes in data distribution over time.

\begin{table}[t!]
    \tabcolsep=0.3cm
    \begin{subtable}{.45\textwidth}
    \centering
    \begin{tabular}{lcccc}
    \toprule
        Model & URLNet & Malconv & LSTM & Total \#\\
    \midrule
        Train & 99.8 & 99.9 & 99.9 & 56754 \\
        Valid & 99.8 & 97.8 & 99.9 & 5248 \\
        Test & 67.8 & 63.5 & 69.1 & 132894 \\
    \bottomrule
    \end{tabular}
    \caption{AUC score when evaluate on open-domain testing set. The ratio of training and validation is 9:1.}
    \label{tab:bad_performace_a}
    \end{subtable}
    
    \begin{subtable}{.45\textwidth}
    \centering
    \begin{tabular}{lcccc}
    \toprule
        Model & URLNet & Malconv & LSTM & Total \#\\
    \midrule
        Train & 99.9 & 99.8 & 99.9 & 63428\\
        Valid & 98.2 & 97.8 & 98.3 & 6344 \\
        Test & 69.3 & 64.1 & 61.6 & 66447\\
    \bottomrule
    \end{tabular}
    \caption{AUC scores when performing backtest on the daily queries dataset. We use 2013-2017 as training and 2018-2019 as testing.}
    \label{tab:bad_performace_b}
    \end{subtable}
    
    \caption{Two experiments that show critical performance drop in close reality setting. Details of the testing set is in section~\ref{sec:dataset}.}
    \label{tab:bad_performace}
\end{table}

Upon closer examination with interpretable machine learning techniques, we identified a severe data bias issue at the heart of this problem. This bias, caused by irrelevant but correlated relations between input tokens and output classes, misled the algorithm into making erroneous predictions.
To tackle this bias issue, we introduce a debiasing feature embedding strategy. Trained in an end-to-end model using adversarial techniques and a blend of classification and debiasing loss, this method aims to eliminate target bias.
Applicable to most end-to-end neural network models, our strategy demonstrated substantial improvements in generalizability (i.e., AUC from 6x\% to 9x\%) when tested on state-of-the-art CNN-based and RNN-based malware detection models with real-world queries.

In summary, this paper makes the following contributions: First, to our understanding, this is the first work that reports severe concern on the generalization of a malware classification model and further argues that we should carefully choose the validation and testing data, as opposed to a random k-fold validation, to prevent from overfitting the bias. We then perform an in-depth analysis to discover that such performance degradation majorly comes from the token-level data bias issue in the security domain. Finally, we propose a debiasing solution based on the adversarial training technique allowing end-to-end classification models to learn a bias-free feature embedding. We demonstrate the proposed strategy can be applied to three state-of-the-art neural-based malware detection frameworks with significant performance boosts.

In the following sections, we will start by introducing the state-of-the-art neural network classification architecture for malicious URL detection in section~\ref{sec:malware}. Next, we will describe how one can interpret such models to identify the data bias problem in section~\ref{sec:investigate}. Finally, we propose an adversarial debiasing architecture in section~\ref{sec:formula} and demonstrate through several experiments how a more robust model can be generated under the proposed framework in section~\ref{sec:exp}.

\begin{figure}[t!]
    \centering
    \includegraphics[scale=0.55]{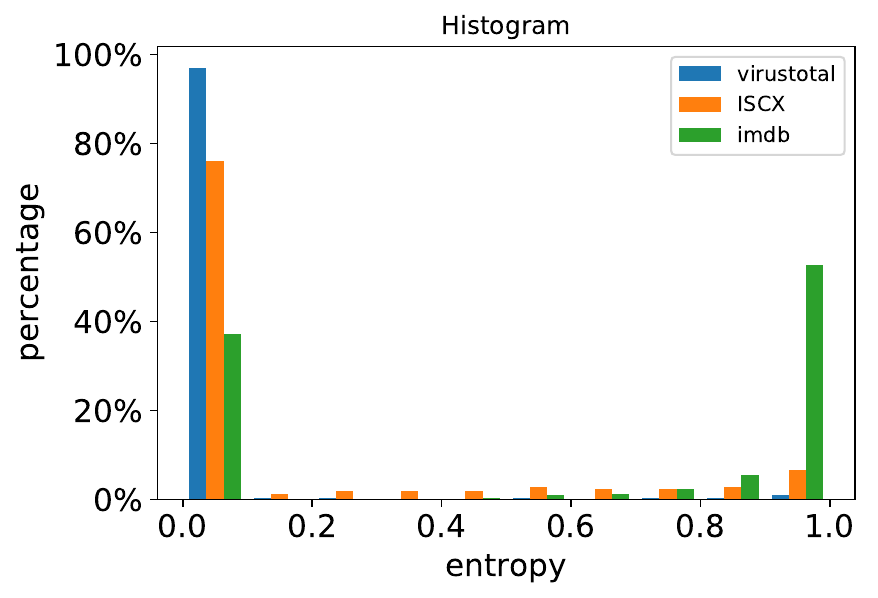}
    \caption{The blue bars represents the distribution of token class distribution entropy in IMDB movie review dataset for binary sentiment classification. The orange bars represents token class distribution entropy in ISCX-URL-2016 URL dataset. The low class-level entropy in URL dataset shows that most token in URL dataset only appears in one specific class.}
    \label{fig:entropy}
\end{figure}

\section{Related Work}
\label{sec:malware}

\subsection{Malware Classification}
\subsubsection{Problem Definition}
Given a malware dataset \(D = \{x_i, y_i\}_{i=1}^{n}\), where each pair \( (x_i, y_i) \) consists of a malware URL file and its corresponding binary label indicating whether it is benign or malicious, our objective is to devise a classifier capable of recognizing malicious files in out-of-domain test data or future instances. While our primary focus lies on URLs, auxiliary experiments detailed in section~\ref{sec:exp} further demonstrate that our framework remains effective for executable and HTML files.

\subsubsection{Model Architectures}
To achieve state-of-the-art performance, end-to-end malware classification operates without relying on domain knowledge or feature engineering. Typically, a sequence of tokens (bytes, words, characters) is converted into embeddings that undergo further classification, akin to sequence or image classification tasks. Below, we introduce three paramount model architectures:

\begin{itemize}
    \item \textbf{LSTM Model:} ~\cite{pascanu2015malware,athiwaratkun2017malware} introduced architectures based on RNN language models augmented with attention mechanisms and max-pooling.

    \item \textbf{MalConv:} In their work, ~\cite{raff2018malware} presented a gated-CNN model that classifies programs based on raw byte sequences without any feature extraction, achieving a cross-validation AUC score exceeding 98\% on their dataset.

    \item \textbf{URLNet:} ~\cite{le2018urlnet} designed URLNet, which utilizes CNN on both characters and words of a URL string to learn the URL embedding in a co-optimized framework, also reporting a cross-validation AUC score above 98\% on their dataset.
\end{itemize}


\subsection{Invariant Feature Learning}
The drive to learn representations invariant to specific factors has inspired research across machine learning fairness, debiasing learning, and domain adaptation, aiming to control the information encapsulated within features.
For instance, studies such as ~\cite{edwards2015censoring,beutel2017data} deploy adversarial training to cultivate fair representations. Works like ~\cite{louizos2015variational,zafar2015fairness} strive to attain a fair representation by minimizing the correlation between the latent representation and sensitive attributes.
Other efforts, like ~\cite{xie2017controllable}, use domain-adversarial training to craft representations that remain invariant to specific factors, enhancing generalization. Meanwhile, studies like ~\cite{li2018domain,akuzawa2019adversarial} employ domain-adversarial training methodologies to achieve controllable invariant features regarding existing variables for domain generalization.
Additionally, ~\cite{kim2019learning} suggests a regularization method that minimizes the mutual information between feature embedding and bias while training neural networks on biased images. These techniques predominantly leverage an adversarial training approach, aiming to either preserve or exclude certain information from the representation.

\begin{figure*}[t!]
  \centering
    \begin{subfigure}[t]{0.99\textwidth}
    \includegraphics[width=\textwidth]{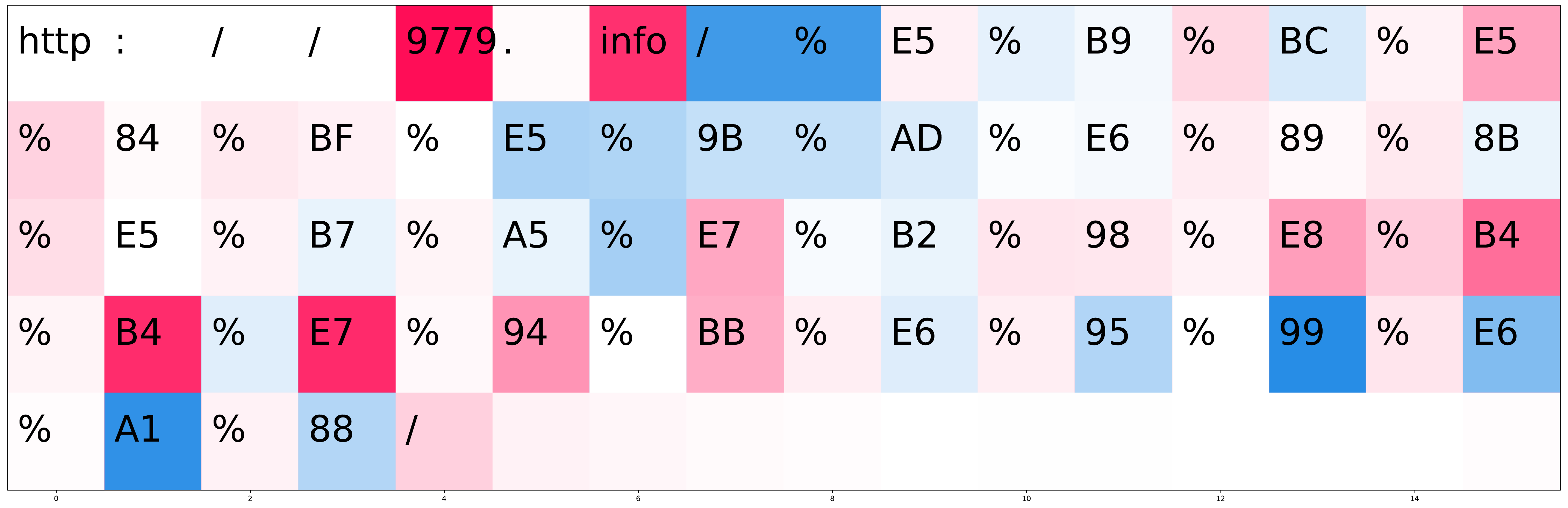}
    \end{subfigure}
    \includegraphics[width=0.99\textwidth]{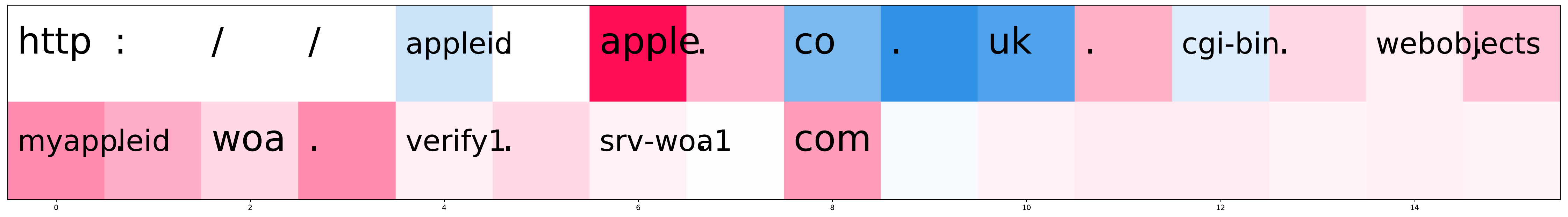}
    \includegraphics[width=0.99\textwidth]{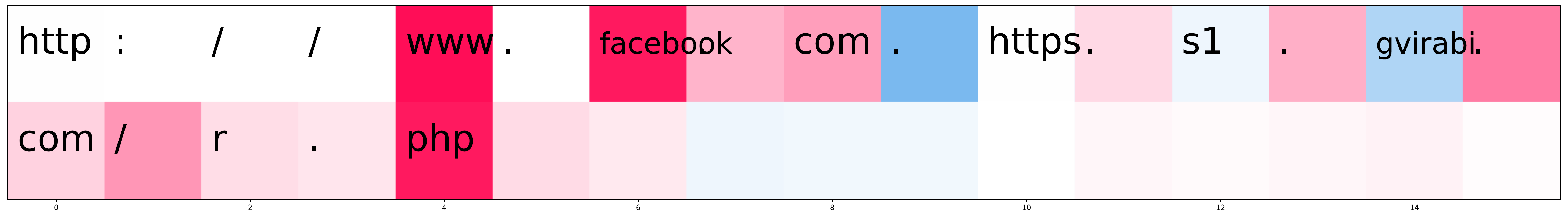}
    \caption{We demonstrate three url samples with the visualized interpretation of URLNet. Red refers to the attribution of the malicious class, and Blue refers to the benign class. The first example shows unreasonable highlights on URL-encoded tokens. The second and third example highlights domain names. The highlights are strongly correlated with the tokens' class-level distribution, which is shown in table~\ref{tab:interpret_ratio}. These three samples can be found on VirusTotal.}
   \label{fig:deep_interpret}
\end{figure*}

\section{Generalizability in Malicious URL Detection Models}
\label{sec:investigate}
In security applications, where misclassification can lead to fatal consequences, ensuring that the trained model accurately learns meaningful patterns for decision-making becomes paramount. This section delves into how techniques from the realm of explainable machine learning can be utilized to discern what types of malicious patterns a classifier identifies. This understanding aids in explaining why a model may fail to generalize to out-of-domain data. We also uncover a significant data bias concern that, to the best of our knowledge, has never been raised in the context of malware detection tasks.

\subsection{Understanding the Malicious URL Classifier}
\label{sec:deep_interpret}
As illustrated in Table 1, we found that the trained classifiers exhibit instability when handling future daily queries. To glean insights into this phenomenon and validate the model's decisions, we specifically interpret URLNet~\cite{le2018urlnet} to elucidate the malicious patterns it identifies. Although the other two models produce similar patterns, we have chosen to omit their detailed analysis due to space constraints.
We utilized the Integrated Gradient algorithm~\cite{sundararajan2017axiomatic} in conjunction with the SmoothGrad algorithm~\cite{smilkov2017smoothgrad} to quantify the attribution of input features to gradients in the form of sensitivity maps. Here, we provide a brief overview of these techniques:
\textbf{Integrated Gradient:} This has emerged as an effective tool to probe the significance of each feature towards predictions in neural networks.
Consider a deep network function \( F: \mathbb{R}^n \rightarrow [0,1] \), and an input \( x = (x_1, \ldots ,x_n) \in \mathbb{R}^n \).
The attribution of the prediction at input \( x \) relative to a baseline \( x^\prime \) is expressed as a vector \( A_F(x, x^\prime) = (a_1, \ldots ,a_n) \in \mathbb{R}^n \), where \( a_i \) represents the contribution of \( x_i \) to the prediction \( F(x) \). Integrated gradients are calculated as the path integral of the gradients along the straight line from the baseline \( x^\prime \) to the input \( x \). The attribution for the \( i^{th} \) feature is given by:
$$(x_i-x^\prime_i)\times\int_{\alpha=0}^{1} \tfrac{\partial F(x^\prime + \alpha\times(x-x^\prime))}{\partial x_i  }~d\alpha$$

\textbf{SmoothGrad:} To mitigate the effect of noisy gradients resulting from essentially inconsequential local variations in partial derivatives, a stochastic approximation of the local average is computed by random sampling in a neighborhood of an input \( x \):
$$\hat{M_c}(x) = \frac{1}{n} \sum_1^n M_c(x + \mathcal{N}(0, \sigma^2))$$
where \( n \) is the number of samples, \( M \) is the sensitivity map, \( c \) is the class of input \( x \), and \( \mathcal{N}(0, \sigma^2) \) symbolizes Gaussian noise with standard deviation \( \sigma \).



\begin{table}[h]
    \centering
    \tabcolsep=0.1cm
    \begin{tabular}{lc|lc|lc|lc}
    \toprule
        9779 & 1.0 & info & 0.93 & www & 0.98 & facebook & 0.89\\
        B4 & 0.76 & E7 & 0.85 & php & 0.81 & apple & 0.97\\
        A1 & 0.28 & 99 & 0.39\\
    \bottomrule
    \end{tabular}
    \caption{The class-level distribution of highlighted tokens in figure~\ref{fig:deep_interpret}. 1.0 means the token only appears in malicious class and 0.0 for benign class.}
    \label{tab:interpret_ratio}
\end{table}

\begin{table*}[th!]
    \centering
    \tabcolsep=0.16cm
    \begin{tabular}{|llcl|lcl|lcl|}
    \toprule
        Year & 2013-2015 & ~ & 2016 & 2015-2017 & ~ & 2018 & 2016-2018 & ~ & 2019\\
    \midrule
        AUC & train 89\% &~& test 52\% & train 91\% &~& test 51\% & train 86\% &~& test 53\%\\
    \toprule
        \multicolumn{10}{|c|}{VirusTotal top 5 token} \\
    \midrule
        & 8i8bfc        &~& 26topmargin     & 26topmargin      &~& Kansascity        & bernice                  &~& bostondyn  \\
        & danlas        &~& poefh3dpzd      & ye               &~& 881633             & tutor-profile            &~& marinadesign  \\
        & cars          &~& Wbot            & poefh3dpzd       &~& sharitzgroup       & serious-70s-scuzz        &~& reachoutmarketing  \\
        & boy           &~& amtrakphotos    & Wbot             &~& bernice            & byfirst                  &~& gv  \\
        & aspnet        &~& 253DYes         & hopefellowshipme &~& heidman            & 882868                   &~& valtho  \\
    \midrule
        \multicolumn{10}{|c|}{ISCX top 5 token} \\
    \midrule
        & E3        &~& \&              & E3     &~& net        & login     &~& 9779  \\
        & 83s       &~& naylorantiques  & style  &~& login      & battle    &~& olx  \\
        & hc360     &~& ;               & hc360  &~& en         & en        &~& post  \\
        & 2015      &~& amp             & detail &~& battle     & ;         &~& distractify  \\
        & torrent   &~& net             & 2015   &~& ;          & amp       &~& back  \\

    \bottomrule
    
    
    \end{tabular}
    \caption{
    The top 5 least diverse tokens, with the smallest entropy in class distribution, reveal the substantial disparity between the training set and testing set. The datasets used include the ISCX dataset and daily queries collected from VirusTotal from 2013-2019. We create a three-fold dataset using a sliding window over time, illustrating the rapid changes in token distribution. Despite selecting a well-fitted model based on the validation set, the AUC score in the testing set on the VirusTotal dataset exhibits a significant drop.}
    \label{tab:topk}
\end{table*}

\subsection{Discovering Data Bias}
There are several established patterns used to identify malicious URLs, such as random domain names (DGA algorithm), suspicious paths (LFI), embedded URLs (SSRF, XSS), code injections, and file downloading. However, upon analyzing the interpretation from URLNet, as shown in Figure~\ref{fig:deep_interpret}, we made a counter-intuitive observation: the highlighted patterns considered critical by the model do not necessarily align with human knowledge of typical malicious URLs. Surprisingly, many highlighted tokens, such as domain names like \textit{facebook} and \textit{apple}, are not usually associated with malicious behavior. Further investigation revealed that these seemingly non-malicious tokens were present in certain phishing URLs like http://www.facebook.com.https.s1.gvirabi.com/r.php and http://appleid.apple.co.uk.cgi-bin.webobjects.myappleid.woa.verify1.srv-woa1.com, causing them to be misclassified as positive instances during training. Those tokens have never appeared in the benign class, which makes them perfect features to be used by an ML model. Additionally, a sequence of meaningless URL-encoded tokens like \textit{\%B4, \%E7, \%99} were highlighted, raising concerns about their significance in the decision-making process from a security expert's perspective.

\begin{figure}[b!]
    \centering
    \includegraphics[scale=0.19]{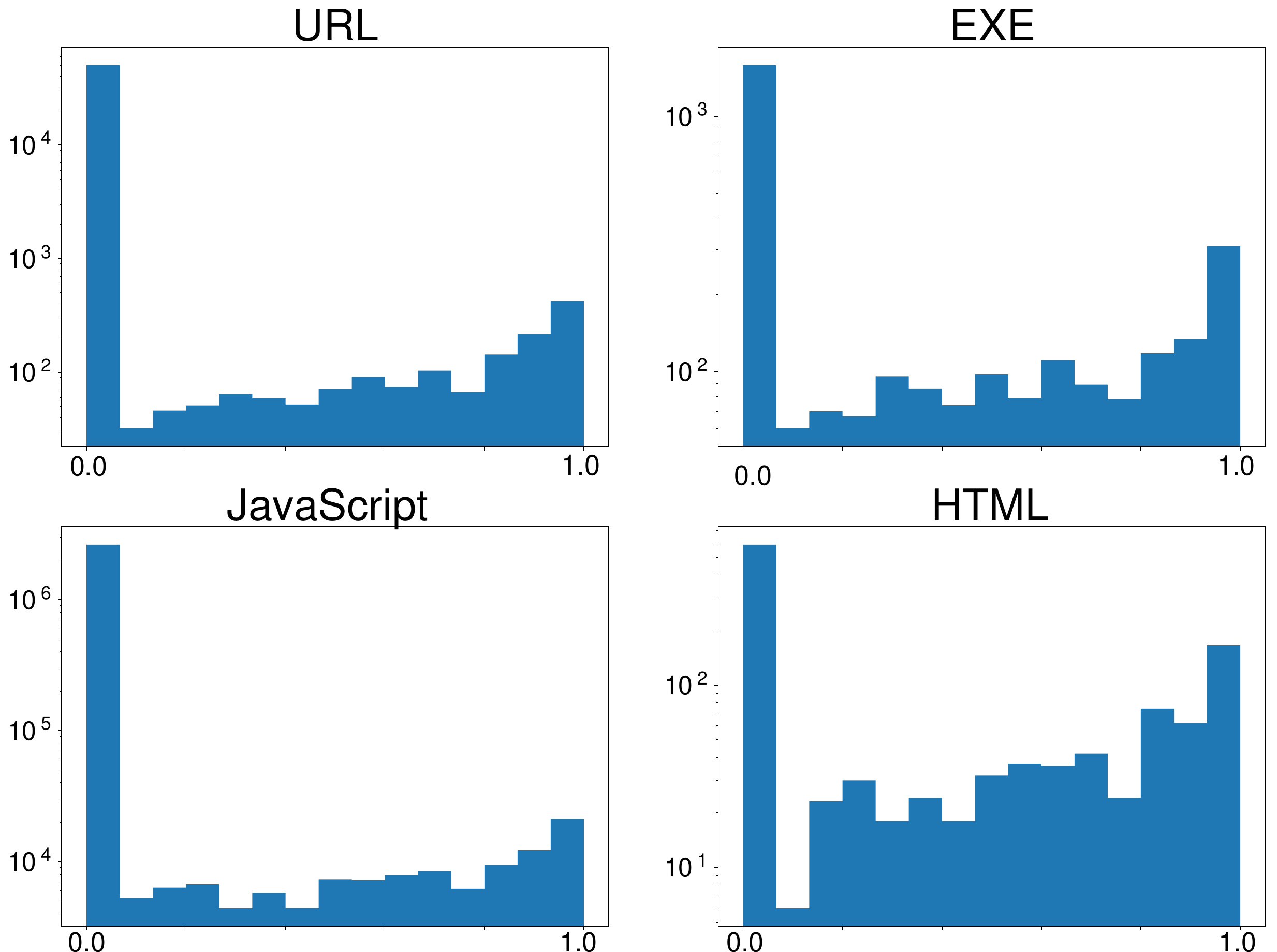}
    \caption{We calculate the entropy of class distribution for each token that appears in the malicious class and the benign class. $1.0$ means the token only appears uniformly in different class and $0.0$ means the token only appears in certain class. The histogram shows severe data bias in all four categories.}
    \label{fig:ratio}
\end{figure}

To gain a better understanding, we conducted a thorough examination and introduced the concept of \emph{class-level distribution} for tokens, representing their occurrence probabilities across different classes. For example, if a token appears 2 times in the malicious sample and 3 times in benign samples, the class distribution would be [2/5, 3/5]. We calculated the class-level entropy for the highlighted tokens in the malicious class (Table~\ref{tab:interpret_ratio}) and, interestingly, found that these tokens exhibited low entropy, making them crucial and distinguishable features among the classes. Figure~\ref{fig:ratio} illustrates the class-level distribution of the tokens highlighted by the IG method, emphasizing the importance of tokens with extremely high or low distribution.

The significance of low-entropy tokens lies in their association with specific classes; they tend to occur predominantly in either malicious or benign URLs, but not both. This property makes them attractive to machine learning models as representative signals for distinguishing between classes. However, relying heavily on such low-entropy tokens can lead to overfitting issues, especially when they lack generalization across different domains, datasets, or time spans.
In particular, we found that such data bias issue has an even higher tendency to happen in URL or other sources for security applications due to the following three reasons:

\begin{enumerate}
    \item There are generally more low-entropy tokens in URL and malware detection tasks compared to other NLP tasks: Low-entropy tokens refer to those that are less informative to human experts, and experienced experts typically do not heavily rely on them for making judgments. Examples of such tokens include the scheme, path, and domain names in URL components. Despite their seemingly less informative nature, some of these tokens exhibit very low class-level token entropy, meaning they tend to occur predominantly in specific classes, rendering them distinguishable features. Figure~\ref{fig:entropy} illustrates a histogram of class-level token entropy, comparing a malicious URL detection task with a regular NLP text classification task (i.e., sentiment analysis using the IMDB user comment dataset). Notably, URL and malware data contain a significantly larger number of tokens with low class-level entropy compared to typical NLP data. Leveraging these low-entropy tokens, a powerful machine learning model can effectively capture critical signals, leading to high cross-validation accuracy.

    \item Malicious patterns change rapidly: To avoid detection, malicious data usually mutate rapidly, causing the class-level token distributions to vary over time. Table~\ref{tab:topk} shows the more representative tokens (i.e., the tokens with the lowest class-level entropy ) in different time span from 2013 to 2019. It reveals that the useful or representative tokens/features identified in the training data (from T-4 to T-1) vary significantly from the ones for testing data at time T, while such diversity is less obvious in NLP tasks.
    Such concept-drift phenomenon implies that a model cannot rely on the low-entropy tokens as critical features to perform classification though it may seem very effective during training and validation time.

    \item Generating a training dataset that accurately reflects the real-world class distribution poses a significant challenge, particularly when collecting data through security intelligence services like VirusTotal, as it can introduce bias. In real-world scenarios, users typically encounter a far greater number of normal URLs (negative samples) than malicious ones (positive samples). However, in several existing malicious URL datasets, the negative data consists of suspicious URLs that are eventually marked as benign. Consequently, URLs that are obviously benign, such as those belonging to famous companies, are often excluded from negative examples. Even if some benign instances are included, their abundance can lead to an imbalanced dataset with an ultra-sparse representation of positive or malicious instances. For instance, tokens like 'Facebook' are frequently encountered in benign URLs, but in training data (e.g., ISCX and VirusTotal), such benign tokens may be absent, and worse, they could mistakenly appear in the malicious class due to mimicking phishing sites. Therefore, detection models must possess the ability to capture genuine malicious patterns while discerning and disregarding these strong yet biased statistical signals.

\end{enumerate}

The phenomenon discuss above also reveals that to evaluate a malware detection model, simply splitting a given dataset for k-fold cross validation can be misleading. It is required to obtain a separate testing dataset through either different sources or collecting at different time-span to make the dataset more robust as possible.




Data bias can cause classifiers to behave like blacklists, memorizing low-entropy tokens rather than capturing complex malicious patterns. Such classifiers may fail to generalize well to new data, leading to performance degradation over time. This phenomenon is not unique to URL data and can be observed in other categories of malware files, such as HTML, executable binary (EXE), and JavaScript (Figure~\ref{fig:ratio}).

\label{sec:bias}

\begin{figure}[!t]
    \includegraphics[width=.465\textwidth]{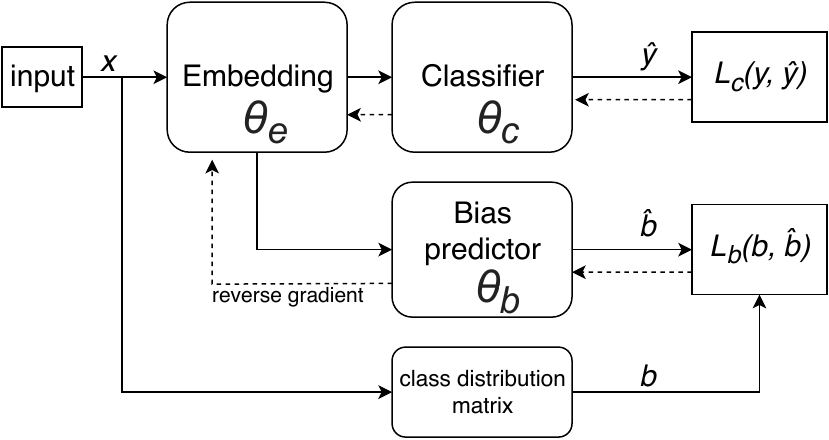}
    \caption{The overall architecture of the adversarial debiasing training procedure.}
    \label{fig:architech}
\end{figure}

\begin{table}[t]
    \centering
    \tabcolsep=0.1cm
    \begin{tabular}{lcccc}
    \toprule
        Category & URL & HTML & EXE\\
    \midrule
        Train & 56754 & 59164 & 25088\\
        Test & 377529 & 6573 & 2787\\
    \bottomrule
    \end{tabular}
    \caption{Number of train, test data within each category.}
    \label{tab:dataset}
\end{table}

\section{Adversarial Debiasing Feature Embedding}
\label{sec:formula}
As we have previously identified, overfitting may occur when relying on individual tokens or features to identify malicious URLs. A straightforward way to alleviate this concern might be to remove individual tokens from the feature set. However, in the context of URL detection, all features are composed of tokens, so simply removing token features is not feasible. To address this challenge, we propose a strategy termed \emph{adversarial debiasing embedding} (ADE), applicable to most end-to-end neural network-based solutions.

The fundamental idea of our proposed method is to eliminate the bias wherein a token is mistakenly correlated with specific classes. Once the biased information is removed, the embedding itself cannot be classified into any particular class. We accomplish this by introducing an adversarial component, called the \emph{Bias Predictor}, to predict the class-level distribution of tokens. We then back-propagate the \emph{reverse gradient} to update the embedding to remove information that is useful for the task of the \emph{Bias Predictor}. Our aim is to minimize the mutual information between discriminating features and biased information so that the learned embedding can serve as inputs to the classifier to learn more complex patterns without relying on biased tokens. The overall architecture is depicted in Figure~\ref{fig:architech}.
Unlike domain adaptation tasks in NLP/CV, which require explicit domain labels, we do not have bias labels for our objective function. A novel aspect of our approach is that we extract biased information directly from the data and use it as pseudo-labels in a manner similar to contrastive self-supervised learning.

To prevent the model from erroneously correlating biased tokens to a specific class, we define the class distribution of tokens as the biases that must be removed from the token embedding. For a $K$-class classification task with a dataset containing $N$ pairs of training data and a vocabulary dictionary $V=\{v_i\}_{i=1}^M$ of $M$ tokens, we calculate a $K \times M$ class distribution matrix $B$, where $b_{ij}$ is the probability of token $v_i$ appearing in class $j$ and vector $b_i$ is a soft class distribution label. Assuming the bias information $B$ is only correlated with the training set:
\begin{align}
    \mathcal{I}(B(x_{train});y) \gg \mathcal{I}(B(x_{test});y) \approx 0,
\end{align}

where $X_{train}$ and $X_{test}$ denote the training and testing datasets respectively, and $\mathcal{I}(\cdot;\cdot)$ denotes the mutual information.

Given an input sequence $x$ and target label $y$, each token in $x$ is transformed into a token embedding after passing through $\theta_e$. The token embeddings are then forwarded to a classifier $\theta_c$ to minimize the classification loss:
\begin{align}
    L_c(x, y; \theta_e, \theta_c) = -\sum_{k=1}^{K}{y_k \log{\hat{y}_k} }
\end{align}
After back-propagating from the classification loss, the token embedding $\theta_e$ learns discriminative features. However, since each individual token in training data are biased, the network $\theta_e$ can rely heavily on the misleading bias information:
\begin{align}
    \mathcal{I}(B(X);\theta_c(\theta_e(x))) \gg 0.
\end{align}

Next, we minimize the mutual information between the token embedding $\theta_e(x)$ and the bias information $B$:
\begin{align}
    \mathcal{I}(B(x);\theta_e(x)) = H(B(x)) - H(B(x)|\theta_e(x))
    \label{equa:MI}
\end{align}

Similar to ~\cite{kim2019learning}, we ignore constant $H(B(x))$ and apply a bias predictor network that approximate the untraceable posterior distribution $P(B(x)|\theta_e(x))$ in eq~\ref{equa:MI} and relax it into minimizing the bias prediction cross-entropy loss which the biases $b$ for each token are obtained from the class distribution matrix $B$. 
When back-propagating, the gradient is passed with gradient reversal technique~\cite{ganin2017domain} in order to eliminate the learned biases in $\theta_e$.
Due to the fact that the bias information vector $b$ is a soft class label, the bias prediction cross-entropy loss can be trained accordingly and characterized as a classification problem. 
\begin{align}
    L_b(x, b; \theta_e, \theta_b) = -\sum\limits_{k=1}^{K}{b_k\log{\hat{b}_k} }
\end{align}

Together with the classification loss and bias prediction loss, the method minimizes the joint minimax loss function where $\lambda$ is the hyperparameter that balances the strength of the bias constraint for the overall objective:

\begin{equation}
    \min_{\theta_e, \theta_c} \max_{\theta_b} L_c (x, y; \theta_e, \theta_c) - \lambda L_b(x, b; \theta_e, \theta_b).
\end{equation}

In each iteration of adversarial training, we first back-propagate the $L_c$ loss to update $\theta_e$ and $\theta_c$. With $\theta_e$ fixed, we then minimize the $L_{b}$ loss to update $\theta_b$. Next, with $\theta_b$ fixed, we maximize the $L_{b}$ loss to update $\theta_e$. After the last step, the bias information is removed from $\theta_e$. Eventually, $\theta_e$ learns feature embedding that is discriminative but does not rely on biased information.

\begin{table}[h!]
    \begin{subtable}{0.47\textwidth}
    \centering
    \tabcolsep=0.4cm
    \begin{tabularx}{1.0\textwidth}{l c c c}
        \textbf{URL} & \textbf{AUC} & \textbf{bACC} & $\textbf{F}_1$\\
        \hline
        URLNet & 69.3 & 67.2 & 68.3\\
        URLNet + BlindEye & 75.1 & 65.5 & 70.3\\
        URLNet + ADE & \textbf{90.1} & \textbf{73.1} & \textbf{70.7}\\
    \bottomrule
    \end{tabularx}
    \caption{}
    \label{tab:url_results}
    \end{subtable}
    \\
    \begin{subtable}{0.47\textwidth}
    \centering
    \tabcolsep=0.4cm
    
    \begin{tabularx}{1.0\textwidth}{l c c c}
        \textbf{URL} & \textbf{AUC} & \textbf{bACC} & $\textbf{F}_1$\\
        \hline
        Malconv & 64.1 & 61.6 & 68.6\\
        Malconv + BlindEye & 72.1 & 65.3 & 69.1\\
        Malconv + ADE & \textbf{90.7} & \textbf{70.7} & \textbf{69.7}\\
    \bottomrule\\
        \textbf{HTML} & \textbf{AUC} & \textbf{bACC} & $\textbf{F}_1$\\
        \hline
        Malconv & 80.7 & 76.2 & 81.6\\
        Malconv + BlindEye & 83.5 & 79.5 & 81.3\\
        Malconv + ADE & 
        \textbf{86.7} & \textbf{81.0} & \textbf{84.3}\\
    \bottomrule\\
        \textbf{EXE} & \textbf{AUC} & \textbf{bACC} & $\textbf{F}_1$\\
        \hline
        Malconv & 67.0 & 67.0 & \textbf{94.7}\\
        Malconv + BlindEye & 68.9 & 70.1 & 93.4\\
        Malconv + ADE & 
        \textbf{71.3} & \textbf{71.2} & 94.0\\
    \bottomrule
    \end{tabularx}
    \caption{}
    \label{tab:cnn_results}
    \end{subtable}
\\
    \begin{subtable}{0.47\textwidth}
    \centering
    \tabcolsep=0.45cm
    \begin{tabularx}{1.0\textwidth}{l c c c}
        \textbf{URL} & \textbf{AUC} & \textbf{bACC} & $\textbf{F}_1$\\
        \hline
        LSTM & 61.6 & 54.0 & 56.4\\
        LSTM + BlindEye & 70.9 & 64.2 & 72.5 \\
        LSTM + ADE & \textbf{90.6} & \textbf{75.5} & \textbf{75.0}\\
    \bottomrule\\
        \textbf{HTML} & \textbf{AUC} & \textbf{bACC} & $\textbf{F}_1$\\
        \hline
        LSTM & 80.8 & 59.4 & 73.6 \\
        LSTM + BlindEye & 84.1 & 80.2 & 82.1\\
        LSTM + ADE & 
        \textbf{87.9} & \textbf{81.2} & \textbf{84.8}\\
    \bottomrule\\
        \textbf{EXE} & \textbf{AUC} & \textbf{bACC} & $\textbf{F}_1$\\
        \hline
        LSTM & 62.5 & 62.5 & 85.5\\
        LSTM + BlindEye & 68.3 & 65.6 & 91.2\\
        LSTM + ADE & 
        \textbf{68.6} & \textbf{68.6} & \textbf{92.3}\\
    \bottomrule
    \end{tabularx}
    \caption{}
    \label{tab:rnn_results}
    \end{subtable}
    \caption{Performance of plain backbone models and backbone+ BlineEye or backbone+ADE. The models are trained using ISCX-URL-2016 dataset and tested using VirusTotal daily queries. The best results in each column are typeset in bold. 
    }
    \label{tab:results}
\end{table}

\begin{figure*}[t!]
    \centering
    \includegraphics[width=0.95\textwidth]{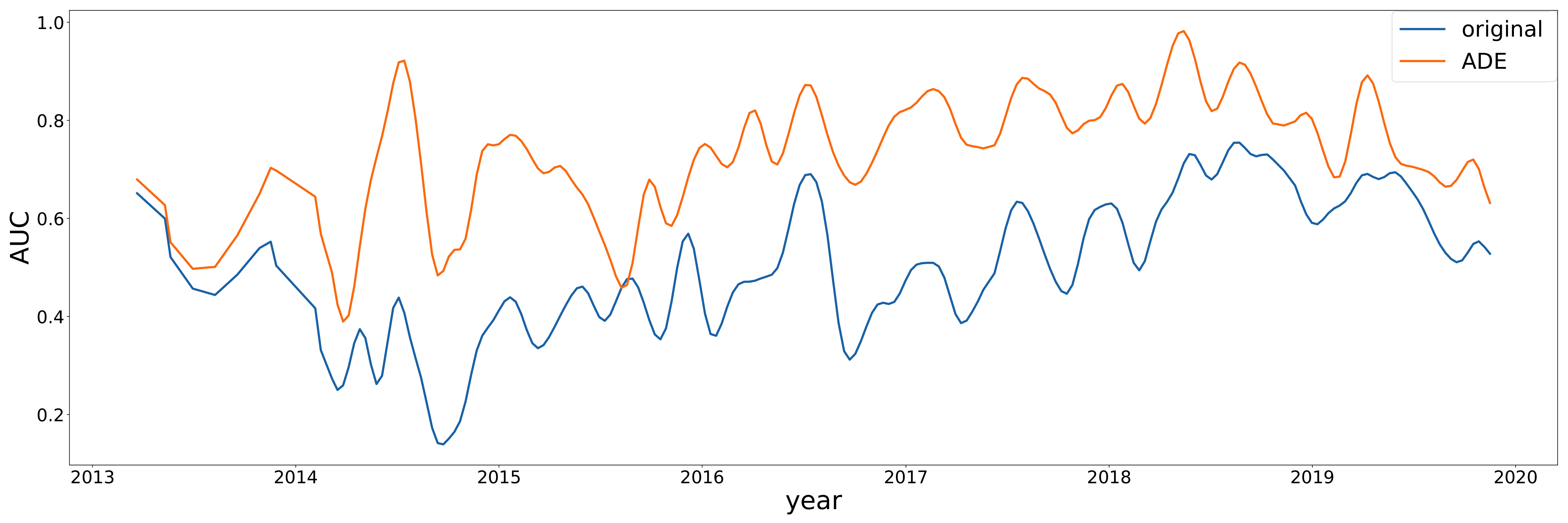}
    \caption{The performance of URLNet trained on VirusTotal over time according to a sliding window on the timeline. URLNet with ADE surpasses the other by a large margin at nearly all times.}
    \label{fig:timeline}
\end{figure*}

\section{Experiments}
\label{sec:exp}
Here we conduct experiments to evaluate whether ADE could lead to better generalization. We evaluate two scenarios: first, whether the proposed solution yields significant improvement over the out-of-distribution data, and second, whether the solution alleviates the degradation of the backbone classifier over time. In addition, we compared the state-of-the-art debiasing feature learning algorithm with the proposed solution.
Note that, as described previously, the majority of the previous experiments divide training and testing data randomly, which are not suitable for evaluating the scenarios of out-of-domain data or data-distribution drift over time. Hence, we have designed two experiment scenarios to better test the generation capability of the proposed solution. In the first experiment, we intentionally choose data from other sources as the out-domain testing data. For example, we trained a model using ISCX-URL-2016 dataset and tested the model using VirusTotal daily queries.
In the second experiment, we made the train/test split according to a sliding window on the timeline to evaluate whether the model can adapt to the data distribution changes over time. That is, a model is trained based on the previous records and evaluated on the future records. For example, the model to be tested using 2018 data is trained with data from 2013-2017.

\subsection{Evaluation Metrics and Datasets}
\label{sec:dataset}
The experiment results are measured by three commonly used metrics, including Balanced accuracy, F1-score, and area under curve (AUC). Among them, AUC is commonly accepted as a critical measurement since it is desirable to have the most malicious files ranked highest and lower false positives.

We collect a large dataset of over one million files in three categories: URL, HTML, and EXE. The URL data contain ISCX-URL-2016 dataset and daily queries from VirusTotal\footnote{https://www.virustotal.com}. The HTML dataset is crawled according to the previously collected URL. The EXE data are obtained from VirusTotal.
The statistics of the datasets are shown in Table~\ref{tab:dataset}.


\subsection{Backbone Models}
In our experiments, we chose three state-of-the-art end-to-end malware classification frameworks as our backbone, two based on CNN (i.e., Malconv~\cite{raff2018malware} and URLNet~\cite{le2018urlnet}) and one based on RNN (i.e., LSTM Malware Language Model~\cite{athiwaratkun2017malware}) structures. 
We want to evaluate whether combining these models with ADE a better generalization capability can be observed. According to Figure~\ref{fig:architech}, we can simply replace the classifier with either Malconv, LSTNet, or LSTM to conduct experiments. Note that URLNet is applied to only the URL dataset due to its inefficiency in longer sequences in HTML and EXE, thanks to the adoption of very small CNN kernels. The details of these models are mentioned in section~\ref{sec:malware}.

Here we compare with the most relevant debiasing algorithm BlindEye~\cite{alvi2018turning} since it also removes bias by learning invariant embedding.
BlindEye explicitly removes biases by minimizing the entropy of the output class distribution on the testing set.
The key difference is that the confusion loss in BlindEye minimizes predictive entropy, while ours minimizes mutual information.
It bounds above our objective function so that we can achieve outperforming results.
BlindEye is originally designed for image classification.

\subsection{Implementation Details}
In the experiments, we use URLNet, Malconv, and LSTM Malware Language Model as the backbone model. The parameters are exactly the same as the original papers.
For the bias predictor, we used a two-layer fully connected network with soft-cross-entropy loss with Adam optimizer and learning rate 1/e.

\begin{figure*}[t!]
  \centering
    \includegraphics[width=0.99\textwidth]{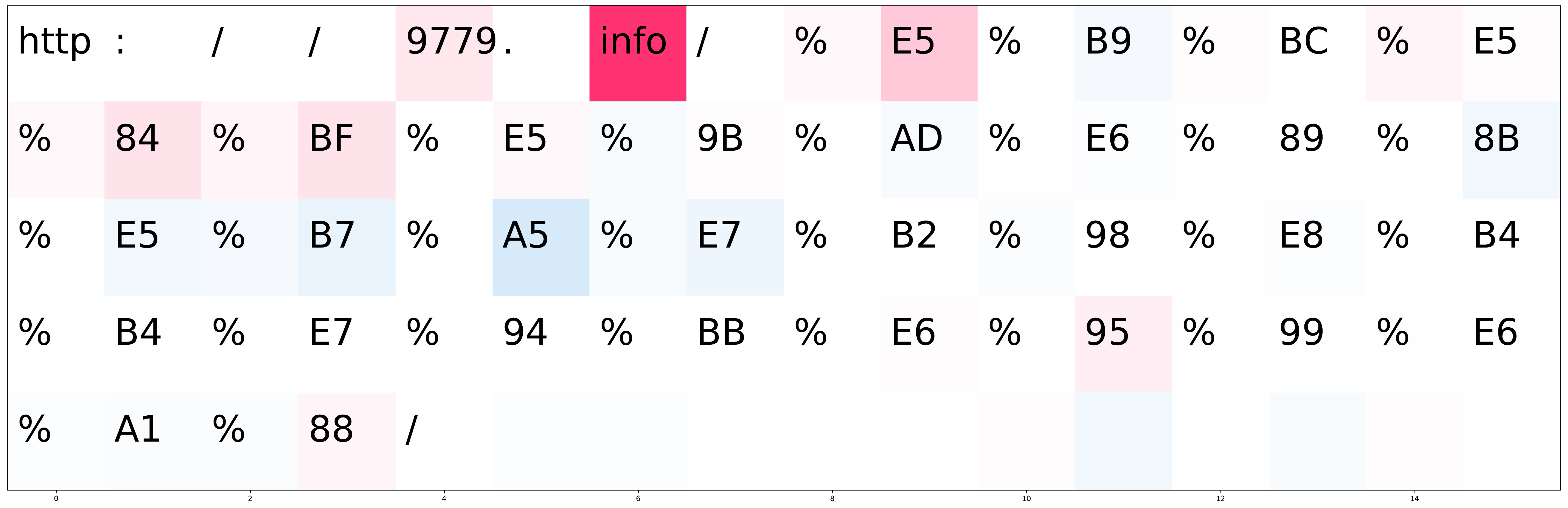}
    \includegraphics[width=0.99\textwidth]{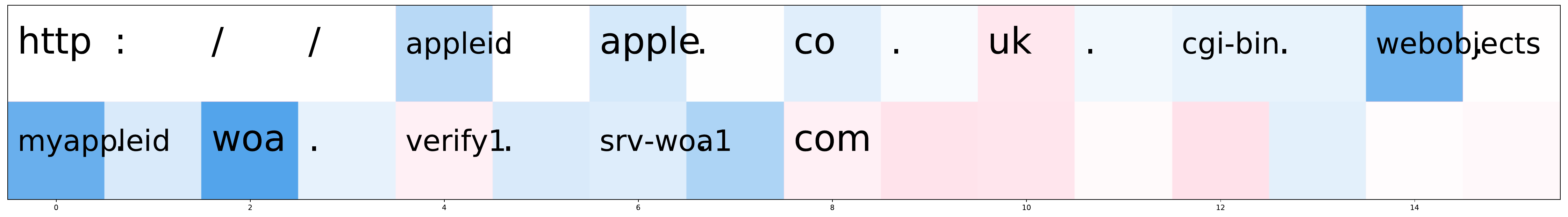}
    \includegraphics[width=0.99\textwidth]{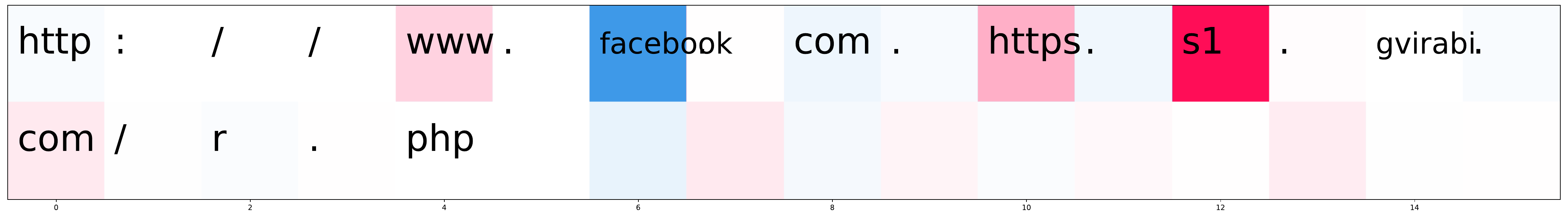}
    \caption{Interpretation for URLNet with debiasing training. The difference with the original one in figure~\ref{fig:deep_interpret} is that domain name such as \textit{facebook}, \textit{apple} turned into negative attribution and url-encoded tokens show nearly no bias.}
  \label{fig:deep_interpret2}
\end{figure*}

\subsection{Results}
We show the performance of the first experiment in Table~\ref{tab:results}. The results show that the proposed algorithm outperformed the original backbone model as well as the BlindEye alternative significantly. The boost in URL data is especially significant since the performance drop in a large held-out testing set is obvious due to data bias and high data variance. Our method improves the AUC score from 67.3\% to 90.6\%.
Data in other categories is much more difficult to collect, which results in a relatively small testing set. This may cause the improvement in generalizability to be less significant, but still, we made improvements on all metrics in Table~\ref{tab:results}. Note that all scores are statistically significant with p-value $<$ 0.001.
Under the scenario of online daily query usage, we trained the model using all data collected before a certain time T and tested the results on data collected in time T. For predicting URLs in time T+1, we retrain the model using all data for the beginning to time T. The results are in Figure~\ref{fig:timeline}. Our proposed method shows obvious improvement in generalizability, which consistently outperform the original model without debiasing. 
In section~\ref{sec:deep_interpret}, we applied URLNet with debiasing embedding and visualized it with an interpretation sensitivity map in Figure~\ref{fig:deep_interpret2}. The interpretation shows a significant difference from the original one in figure~\ref{fig:deep_interpret}. The model no more tries to overfit the performance using domain names caused by data bias.


\section{Conclusion}
In this paper, we report the severe concern about the generalization of a malware classification model. We then perform an in-depth analysis to discover that such performance degradation majorly comes from the token-level data bias issue in the security domain. Finally, we propose a de-biasing solution based on the adversarial training technique allowing end-to-end classification models to learn a bias-free feature embedding. We demonstrate the proposed strategy can be applied to three state-of-the-art neural-based malware detection frameworks with significant performance boosts. We expect this study could be further extended to different usage of data and contributes toward more robust and interpretable ML models for security applications.


\bibliography{aaai24}

\end{document}